\documentclass[letterpaper]{article} 
\usepackage{preprint}
\usepackage[hyphens]{url}  
\usepackage{graphicx} 
\usepackage{natbib}  
\usepackage{caption} 
\usepackage[export]{adjustbox}

\usepackage[inline, shortlabels]{enumitem}

\usepackage{algorithm}
\usepackage{algpseudocode}

\usepackage{xcolor}
\definecolor{BCOBlue}{RGB}{25,90,170}

\newcommand{\bcohl}[1]{{\color{BCOBlue}#1}}
\newcommand{\bcoState}[1]{\State {\color{BCOBlue}#1}}

\usepackage{tcolorbox}
\tcbuselibrary{breakable}
\newtcolorbox[auto counter]{workedexample}[2][]{%
  breakable,
  colback=BCOBlue!5,
  colframe=BCOBlue,
  boxrule=0.5pt,
  arc=1pt,
  left=6pt,
  right=6pt,
  top=4pt,
  bottom=4pt,
  colbacktitle=BCOBlue,
  coltitle=white,
  fonttitle=\bfseries\small,
  title={Example~\thetcbcounter{} (#2)},
  #1
}

\usepackage{microtype}
\usepackage{array}
\usepackage{booktabs}
\usepackage{makecell}
\usepackage{multirow}

\usepackage{listings}

\usepackage{amsmath, amsfonts}

\usepackage[colorlinks=true,linkcolor=blue,citecolor=blue,urlcolor=blue,filecolor=blue]{hyperref}

\usepackage{amssymb}
\usepackage{xspace}

\newcommand{\bco}{\textsc{bco}\xspace}            
\newcommand{\nobco}{\textsc{vanilla}\xspace}      
\newcommand{\BCO}{Belief-Calibrated Optimization\xspace}

\newcommand{\sut}{M}                              
\newcommand{\horizon}{T}                          

\DeclareMathOperator*{\argmax}{arg\,max}

\newcommand{\wmcfile}{\texttt{world\_model\_calibration.md}\xspace}
\newcommand{\predfile}{\texttt{prediction.md}\xspace}

\title{Belief-Calibrated Optimization:\\ An Explicit World Model for Agentic Optimization}

\author{
    Yuhan Chen\textsuperscript{\rm 1},
    Zhihua Tian\textsuperscript{\rm 1},
    Mahavir Dabas\textsuperscript{\rm 1},
    Charith Peris\textsuperscript{\rm 2},
    Rahul Gupta\textsuperscript{\rm 2},\\
    Ming Jin\textsuperscript{\rm 1},
    Feiyang Kang\textsuperscript{\rm 1},
    Siyuan Zhang\textsuperscript{\rm 3},
    Nan Wang\textsuperscript{\rm 3},
    Ruoxi Jia\textsuperscript{\rm 1}
}
\affiliations{
    \textsuperscript{\rm 1}Virginia Tech,
    \textsuperscript{\rm 2}Amazon,
    \textsuperscript{\rm 3}Independent Researcher
}

\copyrighttext{Correspondence to: yuhan@vt.edu}

\begin{document}

\maketitle

\begin{abstract}
The performance of an LLM agent depends on the \emph{scaffold} around
a frozen model. A common way to improve that scaffold is to use a
\emph{coding agent} as an optimizer: it reads current scores and traces
and iteratively edits the source, producing a new candidate each round.
Each edit is chosen according to a belief about how the environment will
respond: what went wrong, and which change should help. That belief is
typically implicit. It lives in the coding agent's reasoning on the current call,
or remains latent in its parameters, rather than as something written down.
Later calls therefore see scores and traces, but they do not use that belief.
We introduce \BCO (\bco), a method that writes that belief down as a persistent
in-context document and continually revises that document as new candidates are
evaluated. The resulting document is a \emph{world model}: the current account
of how the environment responds to edits.
Added to an otherwise standard loop, \bco reaches a higher train passrate than
a matched control that lacks only the world model, on five benchmarks spanning
memory QA, tool-use QA, code-as-action app agents, and terminal agents. The gap
remains on every held-out split, which is not used to select the candidate.
After a target-model swap, in which the frozen model is replaced and
the scaffold is not, the selected \bco scaffold leads on the tasks we test,
except where context-window overruns leave it unfinished. An offline ablation then asks whether that gap comes from
what the world model says. A fresh predictor given the accumulated document
forecasts how the environment will respond more accurately than predictors given
either no document or a same-form copy whose content has been falsified. The
comparison indicates that the document carries reusable information in its
content, not only in its form.
\end{abstract}

\section{Introduction}
\label{sec:intro}

The performance of an LLM agent is determined by both its weights and the
\emph{scaffold} around them. Strong models are typically served as frozen, closed
endpoints, so the scaffold is what a practitioner can actually change. A
recent line of work uses a coding-agent \emph{proposer} as an optimizer: it
reads the current scaffold's scores and traces, diagnoses what went
wrong, and edits the source to produce a new candidate, whose evaluation
outcome then informs the next round~\citep{hu2024adas, zhang2024aflow,yang2023opro,lee2026metaharness,chen2026harnessx,liu2026autoharness,zhang2025dgm}.
Throughout, the frozen model is not updated; only the surrounding program evolves.

Each edit is chosen according to a belief about how the environment will
respond: where the bottleneck is, and which change should help which tasks.
That belief typically remains in the proposer's reasoning or parameters, so
later rounds see scores and traces without using the belief that produced the
last edit.

\begin{figure*}[t]
    \centering
    \includegraphics[width=\textwidth]{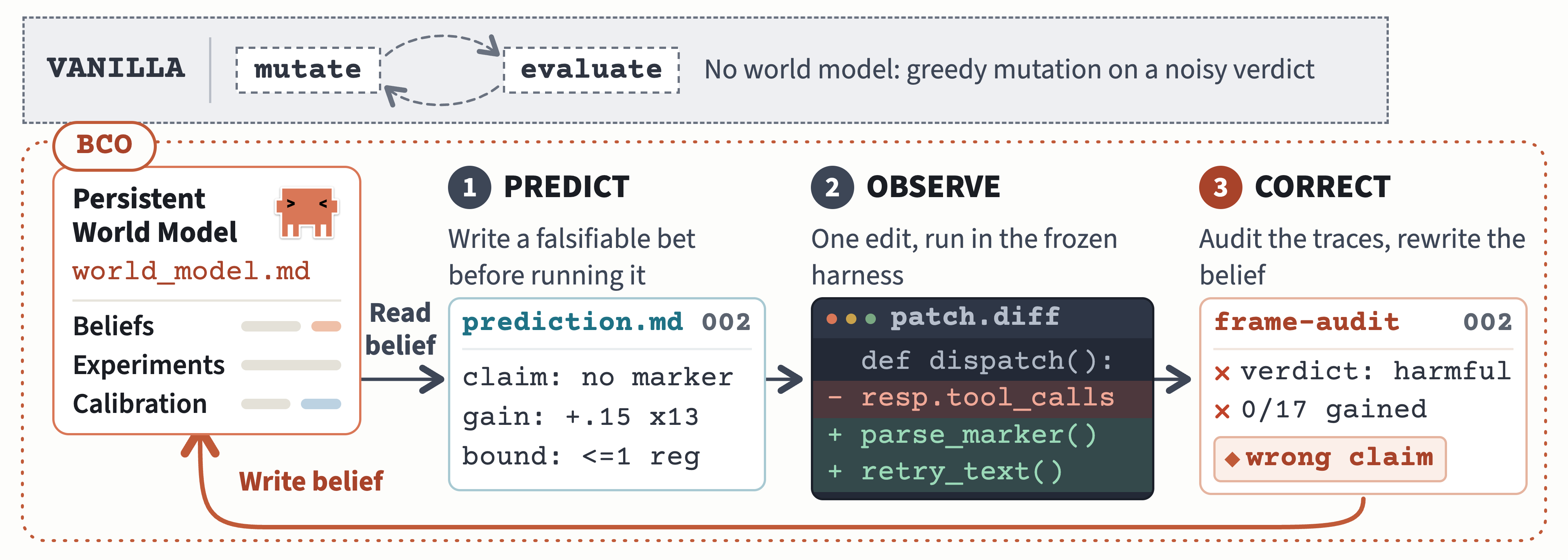}
    \caption{\textbf{Overview of BCO.} \nobco (top) updates the scaffold from
    scores and traces. \bco maintains a persistent world model and repeats three
    steps: predict task-level effects before evaluation (1), apply and evaluate
    a mechanism-level edit (2), and correct beliefs from observed traces (3).}
    \label{fig:teaser}
\end{figure*}

We introduce \BCO (\bco), which writes that belief down as a persistent
in-context document and continually revises that document as new candidates are
evaluated (Figure~\ref{fig:teaser}). The resulting document is a \emph{world model}: the
current account of how the environment responds to edits.
Within a run the target model, tasks, and evaluation stay fixed, so how the
environment responds can be accumulated rather than relearned. This is the
setting of a Bayes-Adaptive MDP, from which we inherit one fact: what to carry
forward is a belief about that
response~\citep{duff2002bamdp,ghavamzadeh2015bayesian}.


Our contributions are as follows:
\begin{enumerate}[leftmargin=1.6em]
  \item \textbf{Belief-Calibrated Optimization.} We present \bco, which
        maintains a world model as a persistent in-context document
        (\S\ref{sec:bco}).
  \item \textbf{Gains across four task families, and evidence for the mechanism
        behind them.} Against a matched control that lacks only the world
        model, \bco reaches a higher train passrate on all five benchmarks, and
        the gap remains on every held-out split. After a target-model swap, the
        selected scaffold leads on the tasks we test, except where
        context-window overruns leave it unfinished. An
        offline ablation indicates that a fresh predictor given the accumulated
        document forecasts environmental responses more accurately than
        predictors given no document or a same-form copy whose content has been
        falsified, so the reusable information is in the content, not only the
        form (\S\ref{sec:experiments}).
\end{enumerate}

\section{Iterative Agentic Optimization}
\label{sec:agentic-opt}

We define \emph{agentic optimization} as an iterative process in which an LLM-based agent
improves an artifact by proposing modifications, evaluating their outcomes, and using the
resulting evidence to guide subsequent proposals. An artifact may be a prompt, program, workflow, or
agent scaffold. At iteration \(t\) the proposer sees the artifacts already evaluated and
chooses one as the parent to edit. An edit \(a_t\) of that parent yields a new candidate
\(x_t\). The environment evaluates \(x_t\) on a fixed \emph{scored} task set \(Q\), drawn once from a task distribution
\(\mathcal{Q}\), and returns an outcome \(o_t\): an aggregate score, task-level results,
execution traces, and tool outputs. After \(T\) iterations the optimizer returns one of the
candidates it actually evaluated, selected by the only utility it can observe:
\begin{equation}
  \hat{x} \;=\; \argmax_{x \in \{x_1,\ldots,x_T\}} \widehat{U}_Q(x),
  \qquad
  \widehat{U}_Q(x) \;=\; U(x) + \varepsilon_x ,
  \label{eq:select}
\end{equation}
where \(U\) denotes the population utility over \(\mathcal{Q}\). In our experiments, it is
estimated on a held-out task set disjoint from \(Q\). The noise term
\(\varepsilon_x\) collects split-specific and evaluation noise. Selecting on
\(\widehat{U}_Q\) while being judged on \(U\) is the optimizer's-curse
setting~\citep{smith2006optimizers}: the candidate that scores highest on \(Q\) is the one
whose \(\varepsilon_x\) is most favourable, so hill-climbing \(\widehat{U}_Q\) alone can lose
utility on \(\mathcal{Q}\). An explicit model of the environment is meant to help the proposer
tell a real effect from a favourable \(\varepsilon_x\).

\section{Belief-Calibrated Optimization}
\label{sec:bco}

\subsection{Overview}
\label{sec:bco_overview}
How the environment will respond to an edit is unknown at the start of a run. Within a run
the frozen target model, the scored set \(Q\), and the evaluation procedure stay fixed, so
each outcome remains evidence about the same response. \bco carries a belief about that
response as a persistent world model \(W_t\). This unknown-but-fixed response, in this loop,
is the setting of a Bayes-Adaptive MDP, from which we inherit the decision to carry a
belief~\citep{duff2002bamdp,ghavamzadeh2015bayesian}. Figure~\ref{fig:teaser} illustrates
the overall workflow of \bco.
At iteration \(t\), the optimizer considers the prior interaction history \(h_t\) (artifacts, edits, scores, and traces so far) and \(W_t\), chooses a parent from those artifacts, and proposes an edit:
\begin{equation}
a_t \sim \pi_{\mathrm{opt}} \left(\cdot\mid h_t,W_t\right).
\label{eq:bco_proposal}
\end{equation}
That edit yields a new candidate. Before evaluation, the optimizer records a
prediction for that candidate at the granularity supported by the protocol
(typically a task subset). The candidate is
then executed to obtain scores and traces.

Before the next iteration begins, \bco compares the predicted and observed candidate outcomes and uses their discrepancies to update the world model:
\begin{equation}
W_{t+1}
=
\operatorname{Update}
\left(
W_t,\widehat{\mathcal{O}}_t,\mathcal{O}_t
\right),
\label{eq:bco_update}
\end{equation}
where \(\widehat{\mathcal{O}}_t\) and \(\mathcal{O}_t\) denote the predicted and observed
outcomes, respectively. The prediction \(\widehat{\mathcal{O}}_t\) is included because it makes the
update local. It specifies which beliefs were staked on the candidate and what outcomes
they implied. The resulting discrepancy can therefore identify the beliefs implicated by the
observed outcome, rather than merely indicating that some revision is necessary. The calibrated
\(W_{t+1}\) is then provided to the optimizer when proposing the next candidate. \bco
thus retains the standard propose-and-evaluate optimization loop while making the beliefs behind
candidate generation explicit and continually revising them against experimental evidence.

\paragraph{Instantiation.}
\label{sec:instantiation}
\(W_t\) is a single Markdown document, \wmcfile, copied into the proposer's workspace at the
start of an iteration and written back at its end. The proposer's instructions are a
\texttt{SKILL.md} file assembled by including other files. One included file, the
\emph{calibration module}, tells the proposer how to maintain that document and how to run
the predict--observe--correct loop.
Removing this include yields \nobco, used as the comparison method throughout our experiments. Both methods receive the \emph{same evidence surface}: every previous
iteration's diff, evaluation output, traces, and full source snapshot, plus the cross-iteration
task--score matrix, all readable on demand. \nobco lacks a persistent place to record its conclusions, so each of its calls must
reconstruct its understanding from the raw evidence.

\subsection{Belief Construction}
\label{sec:belief_construction}

The document \(W_t\) records reusable beliefs formed from the current artifact, task-level scores and
traces, and previous candidate evaluations. We distinguish \emph{execution beliefs} from
\emph{evaluation beliefs} according to which part of the environment's response they describe.


\begin{itemize}[leftmargin=*]
    \item \textbf{Execution beliefs} characterize how an edit is expected to affect agent behavior under particular conditions. They connect properties of the \emph{model, current scaffold, task, and external environment} to anticipated changes in the execution trajectory. For example, they may capture a recurring model limitation, a bottleneck in the current scaffold, a capability required by a class of tasks, or a constraint imposed by an external tool. These beliefs help the optimizer estimate how a candidate edit is likely to affect agent behavior and task performance.
    
    \item \textbf{Evaluation beliefs} characterize how reliably the evaluation system maps an execution trajectory to an observed outcome. They record recurring limitations or errors of the evaluator revealed by previous evaluations, as well as the conditions under which its feedback may be unreliable. These beliefs help the optimizer account for evaluation limitations when comparing predicted and observed performance.
\end{itemize}

We represent \(W_t\) as a collection of atomic beliefs:
\begin{equation}
\begin{split}
W_t &= \{\beta_{t,i}\}_{i=1}^{m_t}, \\
\beta_{t,i}
&=
\left(
\phi_{t,i},
\mathcal{C}_{t,i},
c_{t,i},
s_{t,i},
E_{t,i}^{+},
E_{t,i}^{-},
\mu_{t,i}
\right),
\end{split}
\label{eq:belief_representation}
\end{equation}
where \(\phi_{t,i}\) is a falsifiable claim about the execution or evaluation process,
\(\mathcal{C}_{t,i}\) specifies the conditions under which it applies, \(c_{t,i}\in[0,1]\)
denotes its confidence, \(s_{t,i}\in\{\text{hypothesis},\text{confirmed},\text{refuted}\}\)
its status, \(E_{t,i}^{+}\) and \(E_{t,i}^{-}\) record supporting and contradictory evidence,
and \(\mu_{t,i}\) is the \emph{mass} of the belief: the approximate number of observed failures
it explains. This quantity helps prioritize experimental budget across beliefs. \(m_t\) is the number of beliefs in iteration \(t\). Scores and trace events provide the evidence for these beliefs, while the beliefs themselves capture explanations and generalizations that can inform future candidate generation. \bco retains a task-specific diagnosis in \(W_t\) only when it supports such a reusable claim. In the document these atoms are its \textbf{Beliefs}
section, one entry per \(\beta_{t,i}\), beside an \textbf{Experiments} section recording which
edits already probed each belief and with what verdict, so an edit that paid no
information or gain is not tried again. The document starts \emph{empty}: it is filled only from
the run's own evidence, and pre-seeding guessed failure modes is forbidden, since that would
place confidence on claims no observation supports.

\subsection{Belief Calibration}
\label{ssec:belief_calibration}

After evaluation, the proposer compares the prediction with the observed outcome and updates
\(W_t\) one belief at a time. Beliefs that this outcome does not address are left unchanged.
The update is a short list of operations on \(W_t\):
\begin{equation}
\begin{split}
\Delta W_t &= (\delta_1,\ldots,\delta_{k_t}), \\
\delta_j &\in
\{\textsc{Add},\textsc{Revise},\textsc{Merge},\textsc{Remove}\}, \\
W_{t+1} &= \operatorname{Apply}(W_t,\Delta W_t).
\end{split}
\end{equation}
\textsc{Add} inserts a new belief. \textsc{Revise} changes an existing belief's claim, scope,
confidence, status, or evidence. \textsc{Merge} combines overlapping beliefs.
\textsc{Remove} deletes a belief that the accumulated evidence has refuted.

This update runs at the start of the next iteration, before a new edit is chosen. The proposer
first grades the previous prediction file \predfile against the raw outcome, using per-task
results relative to the parent and the execution traces, not the scaffolded agent's final
message. It
applies \(\Delta W_t\) to the Beliefs section and appends one record to an append-only History,
so a refuted belief remains on the record. It then chooses the next edit: a probe of an
uncertain belief, or an action on a well-supported one. Only then does it write the next
\predfile, before any source change.

The prediction names the belief at stake, the mechanism of the edit, and a claim about a
\emph{subset} of tasks. The candidate code remains general: a prediction may name task IDs, but
the source may not branch on them.

Example~\ref{ex:gaia-case} instantiates this loop on GAIA;
Listings~\ref{lst:pred}--\ref{lst:grade} give the same files verbatim.

\begin{workedexample}[label=ex:gaia-case]{GAIA, iterations 15--16}
\small
\textbf{Predict.}
At iteration~15, the world model identifies buried evidence in long interaction
histories. It predicts that answering from a compressed evidence summary will
improve that long-context subset by at least $0.10$.

\textbf{Observe.}
That summary recovers five tasks but causes four regressions: it
runs after every episode and replaces main-loop answers that were already
correct.

\textbf{Correct.}
Iteration~16 runs the summary turn only when the main-loop answer is empty or a
placeholder, and accepts only a concrete output. Five more tasks pass, none
regress, and train passrate rises from $0.450$ to $0.575$.
\end{workedexample}

Algorithm~\ref{alg:bco} in Appendix~\ref{sec:app-algorithm} states the complete
procedure, marking what \bco adds relative to \nobco. Appendix~\ref{sec:app-protocol} gives the
fragment sizes, the grading axes, and one instance of each artifact.

\section{Experiments}
\label{sec:experiments}

We organize the evaluation around three questions. \textbf{RQ1} asks whether
\bco discovers stronger scaffolds and whether their gains persist on held-out
tasks (\S\ref{sec:exp-main}). \textbf{RQ2} asks how the resulting scaffolds
behave when the frozen target model is replaced by an unseen model
(\S\ref{sec:exp-transfer}). \textbf{RQ3} asks whether the accumulated world
model contains reusable information about how scaffold edits affect task
behavior (\S\ref{sec:exp-ablation}). We study these questions with matched
optimization runs across five benchmarks, target-model swaps, and a controlled
offline prediction ablation.

\subsection{Experimental Setup}
\label{sec:exp-setup}

\paragraph{Matched comparison.} For each benchmark, the two methods
share the initial scaffold $x_0$, the editable source surface, the frozen target
$\sut$ (the open-weight model DeepSeek-V4-Flash on the memory, tool-use, and
code benchmarks, and MiniMax-M3 on Terminal-Bench~2.0; frozen within every
matched pair), the train/test split, the
evaluation harness, and the iteration budget $\horizon$. Both methods start from
the same iteration-0 artifact; the comparison measures the protocol-level
effect of adding the calibration layer in \S\ref{sec:instantiation}.
The proposer is a fixed strong coding LLM at maximum
reasoning effort and runs in a sandbox\footnote{Concretely, the memory-QA
runs use Kimi-K2.6, GAIA and AppWorld Kimi-K2.7, and the Terminal-Bench~2.0
pair a Codex proposer (GPT-5.6 at maximum reasoning effort).}; it is identical
across the two methods within each matched pair.

\begin{figure*}[!t]
  \centering
  \includegraphics[width=\textwidth]{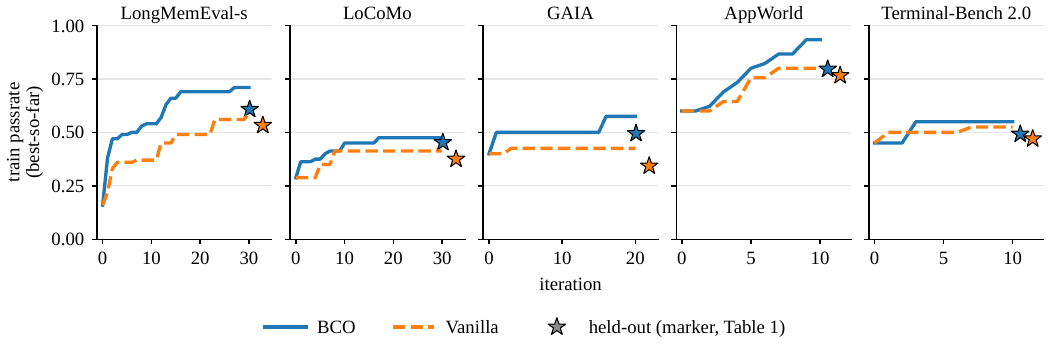}
  \caption{Best-so-far train passrate by iteration (lines); stars mark the
  held-out passrate of each selected candidate. \bco finishes higher than
  \nobco on both train and held-out across all five benchmarks.}
  \label{fig:traincurve}
\end{figure*}

\paragraph{Benchmarks and splits.} We evaluate on five benchmarks spanning
four task families.
\begin{itemize}[leftmargin=1.4em]
  \item \textbf{Memory QA}: \textbf{LongMemEval-s} (LME)~\citep{wu2025longmemeval} (train 100 / test 400) and
        \textbf{LoCoMo}~\citep{maharana2024locomo} (train 80 / test 1449) test
        memory across long, multi-session conversations, including temporal
        and causal reasoning; the scaffold is a long-term-memory ontology with
        retrieval and context packing.
  \item \textbf{Tool-use QA}: \textbf{GAIA}~\citep{mialon2024gaia} (train 40 / test 99)
        poses real-world questions requiring reasoning, browsing, multimodal
        understanding, and tool use; the scaffold is a function-calling loop
        over real tools.
  \item \textbf{Code-as-action agents}: \textbf{AppWorld}~\citep{trivedi2024appworld} (a scenario-disjoint
        train 45 / test 372 split drawn from the challenge pool, so no test
        scenario is seen in training); the scaffold is a ReAct code agent that
        completes tasks in simulated everyday apps by writing Python against
        the environment's APIs.
  \item \textbf{Terminal agents}: \textbf{Terminal-Bench~2.0} (TB2.0)~\citep{tbench2026} (train 20 / test
        66) contains realistic, execution-verified command-line tasks; the
        scaffold is a terminal-use agent loop (Terminus-2) driving a live shell
        in a sandboxed container.
\end{itemize}
Because train score guides the search, we apply a pre-optimization stability
screen before fixing the AppWorld and Terminal-Bench~2.0 train subsets. The
screen removes \emph{oscillating} tasks whose repeated trials disagree under an
identical scaffold, providing a more consistent optimization signal.
Appendix~\ref{sec:app-stability} documents the screen, the Spider2-lite diagnostic, and results on the
initial Terminal-Bench split.

\paragraph{Graders.} Each benchmark keeps its own standard grader, fixed across methods and
identical on train and held-out; Appendix~\ref{sec:app-stability} lists them. Candidates are
evaluated at temperature~0, one repeat per task except Terminal-Bench~2.0 ($k{=}2$).

\paragraph{Metric and candidate selection.} The reported number is held-out passrate. On
GAIA, AppWorld and Terminal-Bench~2.0 each method contributes its top-1 train-frontier candidate,
ties broken by the earliest iteration. Both memory benchmarks run 30 iterations, and their
large headroom over the initial scaffold leaves a group of candidates bunched at the top of the train
split; there each method's \emph{three} highest-train candidates are evaluated held-out and the
best is reported. That procedure selects partly on the held-out
split, so the two memory rows report best-of-three estimates. The same selection
rule is applied to both methods, and
Appendix~\ref{sec:app-stability} lists every evaluated candidate with its train and held-out
score.
Appendix~\ref{sec:app-notes-ablation} provides a complementary three-method diagnostic with free-form
persistent notes, and Appendix~\ref{sec:app-cost} reports proposer token usage.

\subsection{RQ1: Does BCO Improve Optimization and Held-Out Performance?}
\label{sec:exp-main}

This experiment measures both search progress and generalization. The
best-so-far train passrate shows how far each optimizer moves the scaffold on
the split that guides its search; the held-out passrate shows whether the
selected scaffold retains that improvement on unseen tasks. We compare \bco
with \nobco under the matched protocol of \S\ref{sec:exp-setup} across all five
benchmarks.

\begin{table*}[t]
  \centering
  \small
  \setlength{\tabcolsep}{2pt}
  \caption{Held-out generalization across five benchmarks. Each method uses the same
  initial scaffold, editable source, train/test split, target model, proposer,
  evaluation harness, and optimization budget. Train and held-out entries are
  passrates; the held-out split is never used to optimize the scaffold except
  for the disclosed best-of-three selection on the two memory benchmarks.}
  \label{tab:heldout}
  \begin{tabular*}{\textwidth}{@{\extracolsep{\fill}}llccc ccc ccc@{}}
    \toprule
    & & & & & \multicolumn{3}{c}{Train} & \multicolumn{3}{c}{Held-out} \\
    \cmidrule(lr){6-8}\cmidrule(lr){9-11}
    Benchmark & Scaffold & Train/Test & Target & Proposer
      & Initial & \nobco & \bco & Initial & \nobco & \bco \\
    \midrule
    LME & Memory retrieval & 100/400
      & \makecell[l]{DeepSeek-\\V4-Flash} & Kimi-K2.6
      & 0.160 & 0.590 & \textbf{0.710} & 0.148 & 0.533 & \textbf{0.608} \\
    LoCoMo & Memory retrieval & 80/1449
      & \makecell[l]{DeepSeek-\\V4-Flash} & Kimi-K2.6
      & 0.288 & 0.412 & \textbf{0.475} & 0.295 & 0.375 & \textbf{0.453} \\
    GAIA & Tool-calling & 40/99
      & \makecell[l]{DeepSeek-\\V4-Flash} & Kimi-K2.7
      & 0.400 & 0.425 & \textbf{0.575} & 0.283 & 0.343 & \textbf{0.495} \\
    AppWorld & ReAct code & 45/372
      & \makecell[l]{DeepSeek-\\V4-Flash} & Kimi-K2.7
      & 0.600 & 0.800 & \textbf{0.933} & 0.694 & 0.766 & \textbf{0.796} \\
    TB2.0 & Terminus-2 & 20/66
      & MiniMax-M3 & \makecell[l]{Codex\\(GPT-5.6)}
      & 0.450 & 0.525 & \textbf{0.550} & 0.477 & 0.470 & \textbf{0.492} \\
    \bottomrule
  \end{tabular*}
\end{table*}

\paragraph{Optimization result.} Both methods raise train passrate over the
initial scaffold. On every benchmark, \bco ends higher than \nobco (left half of
Table~\ref{tab:heldout}), from $+0.025$ on TB2.0 to $+0.150$ on GAIA.
Figure~\ref{fig:traincurve} shows the best-so-far trajectory.

\paragraph{Held-out result.} Each method's selected candidate is then evaluated
on the split unused for search (right half of Table~\ref{tab:heldout}).
Held-out passrate is lower than train on every row. The ordering is unchanged:
\bco remains higher than \nobco, from $+0.022$ (TB2.0) to $+0.152$ (GAIA).

\subsection{RQ2: How Do the Scaffolds Transfer to an Unseen Target Model?}
\label{sec:exp-transfer}

Scaffold edits can depend on the target used during optimization. We therefore
hold each selected artifact and the held-out tasks fixed, and replace only the
target with a model from a different family and provider. The comparison
separates scores under the original target from scores after that swap.

\begin{figure}[t]
  \centering
  \includegraphics[width=\columnwidth]{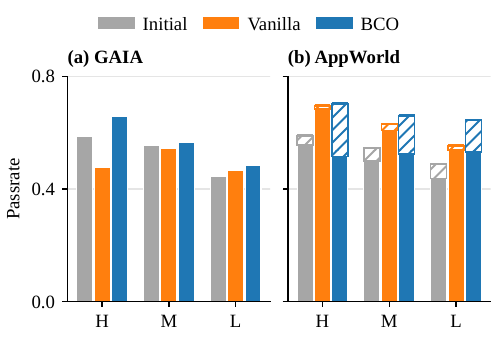}
  \caption{Target-model swap to gpt-5.6-luna at three reasoning-effort tiers
  (H/M/L). Solid bars include all tasks and count non-completions as failures;
  hatched extensions restrict to tasks completed by all three scaffolds. See Appendix~\ref{sec:app-transfer-full} (Table~\ref{tab:transfer-full})
  for exact values and completion counts.}
  \label{fig:transfer}
\end{figure}

\paragraph{Protocol.} For GAIA and AppWorld, we re-evaluate the initial artifact
and the \nobco and \bco selected candidates from Table~\ref{tab:heldout}, using
each run's original source snapshot. The new target is gpt-5.6-luna, at each of
its three reasoning-effort tiers (high, medium, and low). On GAIA the agent loop
is capped at 30 iterations for all three scaffolds, because the swapped target
takes more turns per task.
Figure~\ref{fig:transfer} reports two sets: the full split, counting a
non-completion as a failure, and the tasks that all three scaffolds complete.
Exact values are in Appendix~\ref{sec:app-transfer-full}
(Table~\ref{tab:transfer-full}).

\paragraph{GAIA.} All three scaffolds complete every task, so the two sets
coincide. \bco has the highest passrate at high, medium, and low effort
(Figure~\ref{fig:transfer}). At high effort, \nobco scored $0.111$ below the
initial scaffold. The traces show longer searches that overwrite answers found
earlier in the trajectory; this study does not isolate that mechanism.

\paragraph{AppWorld.} The two sets diverge. The \bco scaffold re-injects its
entire natural-language scratchpad at every turn, with no length bound. That
design stays within the context budget of the original, terser target, but not
of the more verbose swapped target: the scaffold returns no output on $99/77/65$ of the
$372$ tasks at high/medium/low effort. The initial and \nobco scaffolds complete
every task; \nobco keeps a sliding window of $16$ recent exchanges.
Counting non-completions as failures, \bco scores $0.516$, $0.524$, and $0.532$
versus $0.685$, $0.608$, and $0.538$ for \nobco, so \bco is last at high and
medium effort and matches \nobco at low effort. Restricting to tasks all three
complete, \bco is first at every tier, by $+0.007$, $+0.031$, and $+0.091$.
The completed-run ordering is consistent with a transferable task strategy; the
full-set ordering shows that transfer also depends on context consumption.

\subsection{RQ3: Does the World Model Carry Predictive Information?}
\label{sec:exp-ablation}

Train and held-out scores measure the scaffold, not what the world model
contains. We therefore freeze each candidate and its realized outcome, and vary
only the document given to a predictor that did not design the candidate. The
predictor forecasts, for each question type, whether that candidate's edit will
help (\textbf{Upside}) or hurt (\textbf{Downside}).

\begin{table}[t]
  \centering
  {\small
  \setlength{\tabcolsep}{3pt}
  \renewcommand{\arraystretch}{1.15}
  \begin{tabular}{@{}lccc@{}}
    \toprule
    World model & None & Scrambled & Intact \\
    \midrule
    Upside hit rate            & 0.441 & 0.460 & \textbf{0.538} \\
    Downside precision         & 0.142 & 0.181 & \textbf{0.217} \\
    \midrule
    Judge wins vs.\ None      & --- & 26--12 / 24--9 & \textbf{36--4 / 35--2} \\
    Judge wins vs.\ Scrambled & --- & ---            & \textbf{26--12 / 26--12} \\
    \bottomrule
  \end{tabular}}
  \caption{Offline prediction ablation on 40 candidates. Upper rows are
  judge-free; lower rows report blind wins--losses under Kimi-K2.6 /
  DeepSeek-V4-Flash. Ties are excluded. Because candidates are nested within
  four optimization runs, the run is the independent unit; all four runs favor
  the same ordering, giving a two-sided sign-test $p=0.125$ (Appendix~\ref{sec:app-ablation}).}
  \label{tab:ablation}
\end{table}

\paragraph{Protocol.} All predictors receive the same parent source, diff, and
base per-type passrates. \textbf{None} receives no document; \textbf{Intact}
receives the run's final \wmcfile with that candidate's outcome block removed;
\textbf{Scrambled} keeps the file's length, format, and vocabulary but
falsifies its mechanism verdicts and type-level effects. Two blind judges
compare each pair of predictions to the realized per-type changes; upside hit
rate and downside precision are a judge-free cross-check
(Appendix~\ref{sec:app-ablation}).

\paragraph{Result.} Table~\ref{tab:ablation} gives the same ordering on every
measure: None $<$ Scrambled $<$ Intact. Scrambled over None is consistent with a
benefit from structure and from naming regressions. Intact over Scrambled is
consistent with additional information in the learned content. Both judges
agree with the mechanical metrics; the largest mechanical gap is downside
precision. Candidates are nested in four runs, so the run is the independent
unit (Appendix~\ref{sec:app-ablation}; two-sided sign-test $p=0.125$). The
comparison shows that a predictor that did not write the document can use its
content to forecast outcomes of candidates from those runs; it does not test
whether the same document would transfer to a new optimization trajectory.

\subsection{Additional Ablation: Persistence Without Calibration}
\label{sec:exp-notes}

The offline ablation isolates document content. It does not test whether any
persistent file would suffice. On LongMemEval-s we therefore add a third method,
\textbf{Notes}: one freely editable \texttt{agent\_notes.md} that persists
across iterations, with no prediction, grading, or predict--observe--correct
update. \nobco remains the comparison method; \bco is the full protocol.

Three independent ten-iteration trajectories per method on the $100$-example
training split give means $0.430$ (\nobco), $0.427$ (Notes), and $0.467$
(\bco), with within-method ranges $0.11$, $0.07$, and $0.13$
(Table~\ref{tab:notes-main}). Persistence alone does not reproduce \bco's mean
on this diagnostic, but three short trajectories cannot show that free-form
notes are ineffective. The paired held-out comparison and the content-controlled
ablation remain the primary evidence.

\begin{table}[t]
  \centering
  {\small
  \renewcommand{\arraystretch}{1.15}
  \begin{tabular}{@{}lcccc@{}}
    \toprule
    Method & Run 1 & Run 2 & Run 3 & Mean \\
    \midrule
    \nobco & 0.37 & 0.44 & 0.48 & 0.430 \\
    Notes  & \textbf{0.43} & \textbf{0.46} & 0.39 & 0.427 \\
    \bco   & 0.42 & 0.43 & \textbf{0.55} & \textbf{0.467} \\
    \bottomrule
  \end{tabular}}
  \caption{Persistent-state ablation on LongMemEval-s. Entries are the
  best-so-far training passrate over successfully evaluated candidates in
  three independent trajectories (100 training examples; nominal ten-iteration
  proposer budget).}
  \label{tab:notes-main}
\end{table}

\section{Related Work}
\label{sec:related}

\paragraph{LLM-driven optimization of the program around a frozen model.}
Closest to our setting is work that optimizes the \emph{harness}: end-to-end
harness optimization~\citep{lee2026metaharness}, composable and evolvable
foundries~\citep{chen2026harnessx}, sustained self-improvement on open-ended
task streams~\citep{liu2026autoharness}, accelerated self-evolution
loops~\citep{hu2026flashevolve}, and dedicated evaluation
environments~\citep{zheng2026seagym}, extending earlier agents that rewrite
their own scaffold or code~\citep{zhang2025dgm, yin2024godelagent,
shang2024agentsquare, zhou2024symbolic, zweiger2025seal}. A parallel line
optimizes agent systems and prompts: agentic-system design~\citep{hu2024adas},
graph-structured topologies~\citep{zhuge2024gptswarm}, workflow
generation~\citep{zhang2024aflow}, LLM-as-optimizer~\citep{yang2023opro} and
evolutionary~\citep{guo2023evoprompt, fernando2023promptbreeder} prompt search,
and compiling declarative LM programs by optimizing their instructions and
demonstrations~\citep{khattab2023dspy, opsahlong2024mipro}. These methods show
that an LLM can improve the program around a frozen model using observed scores,
often together with free-form reflection, to guide edits over program space. A
concurrent analysis separates producing useful updates from benefiting from them
at solve time and finds update quality roughly flat across base-model tiers.
This pattern suggests that the \emph{update protocol}, rather than the evolver's
raw capability alone, is an important lever~\citep{lin2026harnessupdating}.
\BCO changes what the optimizer carries from one iteration to the next.

\paragraph{Verbal self-reflection and memory.}
Agents can also improve by writing natural-language reflections on their own
behavior---repeated self-feedback on a single output~\citep{shinn2023reflexion,
madaan2023selfrefine}, tool-verified self-correction~\citep{gou2023critic}, a
persistent skill library~\citep{wang2023voyager}, or a streamed memory of
experience~\citep{park2023generativeagents}. These approaches use episodic or
free-form memory to retain useful experience across interactions. \BCO instead
organizes persistent state around explicit predictions, observations, and
revisions of an environment model.

\paragraph{POMDPs, Bayes-adaptive RL, and Bayesian optimization.}
The setting is that of a Bayes-adaptive MDP: the environment's response is
unknown but fixed within a run, so what to carry between iterations is a belief
about that response~\citep{duff2002bamdp,ghavamzadeh2015bayesian,astrom1965,
kaelbling1998pomdp}. Sample-based planners that search the belief-augmented
state space instantiate the same idea in parametric
form~\citep{guez2012bamcp}. The optimization loop can also be read as Bayesian
optimization~\citep{shahriari2016bayesopt}: the world model plays a
surrogate-like role, and the choice between probing an uncertain belief and
acting on a well-supported one plays an acquisition-like role. We instantiate
those roles in natural language and in context, without a parametric surrogate
or a learned solver. That choice follows the view of in-context learning as
implicit Bayesian inference~\citep{xie2021icl}.

\section{Conclusion}
\label{sec:conclusion}

\BCO writes the optimizer's belief about how the environment responds to edits
as a persistent world model, and revises that document by
predict--observe--correct. On five benchmarks spanning memory QA, tool-use QA,
code-as-action app agents, and terminal agents, \bco reaches a higher train
passrate than \nobco and keeps that gap on every held-out split.

When the target is replaced with an unseen model, the \bco scaffold ranks first
across all GAIA settings and on the common completed-task sets in AppWorld. Its
unbounded context use, however, reduces full-set AppWorld performance. A fresh
predictor given the accumulated world model also forecasts outcomes more
accurately, indicating that the document contains information usable beyond the
run that produced it. More broadly, an optimizer's knowledge of its environment
need not remain implicit and disposable. When made explicit, falsifiable, and
calibration-tracked, it becomes an artifact that can be maintained and improved.

\section{Limitations}
\label{sec:limitations}

\begin{itemize}[leftmargin=1.4em]
  \item \textbf{The belief is approximate, and its self-assessment is heuristic.} The world
        model $W_t$ is an LLM's natural-language summary of evidence, not a
        statistically calibrated posterior. Our analysis therefore focuses on de-noised,
        subset-averaged behavior, since single-task pass/fail sits below the evaluation noise floor;
        raising the per-task repeat count would admit stable single-task claims at
        proportionally higher cost.
  \item \textbf{The environment's response is treated as fixed within a run, and the scaffold can overfit the target.}
        We freeze the target model and the benchmark for a run, so the response does not
        drift by design. A target swap or external-tool drift changes the response and would
        require detecting the shift and discounting older evidence, which the
        current protocol does not include. The AppWorld transfer result in
        \S\ref{sec:exp-transfer} also shows that target-dependent resource use can affect
        full-set performance. Future variants could optimize against multiple targets or score
        a candidate's resource envelope alongside its passrate.
  \item \textbf{The evaluation covers five screened benchmarks.} Four benchmarks share one
        proposer family and one target model, while Terminal-Bench~2.0 uses an independent
        stack. Each matched comparison contains one trajectory per method, so the aggregate
        evidence measures consistency across five benchmarks rather than within-benchmark
        repetition; the transfer study changes the target at evaluation time. We also screened
        initial-scaffold behavior before fixing the AppWorld and Terminal-Bench~2.0 train subsets, which
        narrows the claim to environments with sufficient repeatable failure mass and may
        favor benchmarks on which score-guided optimization is easier
        (Appendix~\ref{sec:app-stability}).
\end{itemize}

\clearpage
\bibliography{references}

\appendix
\raggedbottom

\section{The Optimization Loop}
\label{sec:app-algorithm}

Algorithm~\ref{alg:bco} is the procedure of Section~\ref{sec:bco} of the main paper. Every black
line is shared with the matched control; the blue lines are what \bco adds.

\begin{algorithm}[t]
\caption{\textbf{Belief-Calibrated Optimization.}
\bco-specific operations and inputs are shown in blue; every black line is shared with the
matched control, including \textsc{SelectParent}, in which the proposer chooses which previously
evaluated candidate to build on.}
\label{alg:bco}
\begin{algorithmic}[1]
\Require Initial artifact \(x_0\), scored task set \(Q\), and iteration budget \(T\)
\Ensure Selected artifact \(\hat{x}\) and world model \(W\)

\State \(o_0 \gets \Call{Evaluate}{x_0,Q}\);\quad
       \(h_1 \gets (x_0,o_0)\)
\bcoState{\(W \gets
    \Call{ConstructBeliefs}{x_0,o_0}\)}

\For{\(t=1,\ldots,T\)}
    \State \(x^{\mathrm{par}}_t \gets \Call{SelectParent}{h_t}\)
        \Comment{proposer picks any evaluated ancestor}
    \State \(a_t \gets
        \Call{Propose}{x^{\mathrm{par}}_t,h_t,\bcohl{W}}\)

    \bcoState{\(\widehat{\mathcal{O}}_t \gets
        \Call{Predict}{W,a_t,Q}\)}

    \State \(x_t \gets a_t(x^{\mathrm{par}}_t)\);\quad
           \(o_t \gets \Call{Evaluate}{x_t,Q}\)

    \bcoState{\(\Delta W_t \gets
        \Call{Calibrate}
        {W,a_t,\widehat{\mathcal{O}}_t,o_t}\)}
    \bcoState{\(W \gets
        \Call{Apply}{W,\Delta W_t}\)}

    \State \(h_{t+1} \gets h_t \cup (a_t,o_t,x_t)\)
\EndFor

\State \Return \(\hat{x} \gets \argmax_{x\in\{x_1,\ldots,x_T\}} \widehat{U}_Q(x)\),\; \(W\)
\end{algorithmic}
\end{algorithm}

\section{The Protocol, Verbatim}
\label{sec:app-protocol}

This appendix shows the artifacts the main text describes: what separates the two methods, what a
world model and a prediction actually look like, and how the blind judge of
Section~\ref{sec:exp-ablation} of the main paper is configured.

\lstdefinestyle{wmc}{
  basicstyle=\ttfamily\scriptsize,
  breaklines=true,
  breakindent=0pt,
  postbreak=\mbox{$\hookrightarrow$\,},
  columns=fullflexible,
  keepspaces=true,
  language=,
  frame=single,
  framesep=3pt,
  framerule=0.4pt,
  rulecolor=\color{black!25},
  backgroundcolor=\color{black!3},
  xleftmargin=3.4pt,
  xrightmargin=3.4pt,
  aboveskip=6pt,
  belowskip=4pt,
}

\subsection{What separates the two methods}

Each proposer is a \texttt{SKILL.md} assembled by textual inclusion from three fragments. For
the GAIA pair, for instance, both methods include a 61-line environment surface (the editable
source files and the shape of the returned outcome) and a 123-line base contract (read the
evidence, design one mechanism-level edit, write the candidate). The \bco method adds one more
line, \verb|<!-- INCLUDE: _calib_addon.md -->|, pulling in a 169-line calibration module; the
\nobco file is otherwise identical, including the task description. The module is the
only difference between the two methods. It specifies the two regions of \wmcfile and their invariants, the
predict--observe--correct order within an iteration, the grading rule (grade against raw
tool-call outputs and harness control flow, never the agent's own final message; exclude
tasks whose cross-iteration score row disagrees with itself), the four honest causes a missed
prediction can have, and the template for \predfile reproduced below. It imposes no
scheduling rule, no coverage quota, and no list of failure modes to look for.

\subsection{The world model}

Listing~\ref{lst:beliefs} shows the \textsc{head} of \wmcfile from the Terminal-Bench~2.0
\bco run at its final iteration, abridged to three of nine beliefs and three of nine
experiment records. Each belief is one instance of the tuple in
Eq.~\ref{eq:belief_representation} of the main paper: claim $\phi$, confidence $c$, status $s$, evidence
$E^{\pm}$ given as file paths into the stored trajectories, and mass $\mu$. Evidence strings
are elided at \texttt{[...]}; nothing else is edited.

\begin{lstlisting}[style=wmc,caption={\textsc{head} of \wmcfile, Terminal-Bench~2.0, abridged.},label={lst:beliefs},captionpos=t]
## Beliefs
[E3] Byte-exact workspace memory is a material bottleneck when an ordinary stateful
     read destroys the only useful input before the solver recognizes the loss; a
     concrete loss report plus a recoverable pristine object changes behavior and can
     change stable task outcomes
     | conf:0.96 | status:confirmed
     | evidence:base_eval/dumps/iter003_bounded_workspace_journal/db-wal-recovery/
       trajectory.json (steps 3-6: two WORKSPACE_MEMORY_LOSS_DETECTED reports followed
       by exact WAL/SHM restoration); seed_eval/[...] (steps 5-8: SQLite removes the
       only WAL and the solver begins guessing); [...] ctrf.json (7/7 versus seed 5/7)
     | mass:~1

[E4] Comparison-ready artifact history is not a demonstrated bottleneck on this train
     split: when a post-command ledger surfaced recoverable candidates, the solver
     usually continued the same branch without the requested criterion table
     | conf:0.26 | status:refuted
     | evidence:reference_iterations/iter_007/dumps/[...]/largest-eigenval/
       trajectory.json (checkpoint steps 9,18,20,25,28,30 followed by further unranked
       rewrites); [...]; task_score_matrix.json (train-fasttext remains stable 0)
     | mass:~0

[E9] Repeated external failure output is an earlier and more objective stagnation
     signal than plan wording: the same normalized failure can recur across
     substantively different commands while the solver continues modifying parameters
     | conf:0.91 | status:hypothesis
     | evidence:seed_eval/dumps/terminus2_tb2/adaptive-rejection-sampler/
       trajectory.json (steps 19,21,23,24,29,30 repeat "failed to accept a sample after
       100 iterations" across parameter changes and a rewrite); [...]
     | mass:~3

## Experiments
- E3: bounded pre-command content-addressed workspace journal with explicit deletion
      /truncation reports -> held (db-wal-recovery moved stably 0->1, the three-task
      stable subset mean rose +0.333, and no stable task regressed)
- E4: overwritten/deleted-path variant ledger with a third-version checkpoint -> inert
      (the edited source and initial prompt were live, but no checkpoint fired on the
      only evaluated target; source inspection shows the ledger compared only
      pre-existing paths, so the environment claim remains ungraded)
- E6: post-stall action quarantine -> VOID (the iteration-6 score moved, but no
      WORKSPACE_PROGRESS_GATE signature appears in any candidate trace; without a
      candidate-distinctive live signature, those numbers do not update E6)
\end{lstlisting}

Two features are worth noting because they are what the protocol asks for rather than what an
LLM writes by default. \texttt{E4} is retained at \texttt{status:refuted} instead of being
deleted, so the direction is not re-proposed later. The \texttt{E6} entry records a candidate
whose score moved but whose mechanism left no trace signature, and refuses to credit the
belief with the movement.

\subsection{A prediction and its grading}

Listing~\ref{lst:pred} is \predfile from iteration~16 of the GAIA \bco run---the candidate
that became the train-frontier selected candidate and the scaffold behind the GAIA row of
Table~\ref{tab:heldout} in the main text---and Listing~\ref{lst:grade} is the distill block appended at the
start of the next iteration, grading it. Both are abridged; one marker token is rewritten as
\texttt{<|tool\_calls|>} for typesetting.

\begin{lstlisting}[style=wmc,caption={\predfile, GAIA iteration 16, abridged.},label={lst:pred},captionpos=t]
# iter_016 prediction

## Mode
refine -- the `long_context / evidence_noise` cell is open with one partial attempt
(evidence-summary-memory in iter_015). This round refines the same family by gating the
summary turn on placeholder answers and adding a quality guard, which should eliminate
the regressions while keeping the breakthroughs.

## Candidate
conservative_quality_gated_summary -- after the FC loop, run the evidence-summary
final-answer turn only when the current answer is empty, a <|tool_calls|> marker, or a
placeholder; only adopt the summary answer if it is concrete.

## Base
iter_015 -- I copied the iter_015 source snapshot into the editable tree because the
mechanism is a direct refinement of its evidence-summary-memory scaffold.

## Failure mode attacked
`long_context / evidence_noise` (~10-15 train tasks per the map; attempts so far: 1).

## Assumption audit
Not triggered for this cell (<3 attempts). The lower-layer assumptions are: (1) the
scaffold can extract tool-call results from message history, (2) a compact evidence
summary preserves enough information for the model to answer, and (3) a clean summary
context produces a better final answer than a bloated one for episodes whose main loop
emitted a placeholder.

## Mechanism -- the behavioral change
[...] 5. Adopt the summary answer only if it is non-empty AND not itself a placeholder;
otherwise keep `original_answer`. Any exception during the summary turn also falls back
to `original_answer`. [...] It does NOT change behavior on episodes that already
produced a concrete answer, preventing the iter_015 regressions.

## Behavioral prediction (graded from traces)
On the subset, the agent will now keep its concrete direct answers when they exist and
run the clean-context summary turn only when the main loop returns a placeholder/marker.

## Aggregate prediction (graded on the subset, de-noised)
- expected: the subset's mean stable-pass / mean score rises by >=0.10. This is
  plausible if the 5 iter_015 breakthroughs are retained and the 4 regressions are
  eliminated by the placeholder gate.
- downside bound: <=1 currently-stable-passing task regresses, enforced by construction.

## Anchors (optional, STABLE episodes only -- illustration, not the bet)
No stable anchors are available; the relevant episodes oscillate across iterations due
to endpoint stochasticity and marker noise. The mechanism is justified by the general
failure family, not by a single stable episode.

## Falsification
If long-context level-2 episodes that produced placeholders still fail after the summary
turn, or if concrete-answer episodes regress because the summary turn overwrites them,
the mechanism is refuted.
\end{lstlisting}

\begin{lstlisting}[style=wmc,caption={The distill block grading Listing~\ref{lst:pred}, abridged.},label={lst:grade},captionpos=t]
## iter_015 -> iter_016 distill
- Mode: refine
- Behavioral check (self): predicted the agent would gate the evidence-summary turn on
  placeholder/empty/marker answers and preserve concrete direct answers. The mechanism
  is wired correctly: comparing per-episode traces, all 18 tasks that passed in iter_015
  also passed in iter_016 (zero regressions), and the final predictions for
  previously-regressed tasks like gaia#4fc2f1ae and gaia#3f57289b returned to their
  correct direct answers rather than being overwritten by a summary. -> behavior changed
  as predicted on the targeted subset.
- Aggregate check (self): `long_context / evidence_noise` subset (~15-20 episodes) --
  predicted the subset would gain and the 4 iter_015 regressions would be eliminated.
  Actual: 5 additional tasks passed vs iter_015, 0 regressions, overall passrate
  0.45 -> 0.575. Verdict: works.
- Unstable (multi-run, excluded): many tasks have oscillated across prior iterations due
  to marker stochasticity. The zero-regression result vs iter_015 is treated as the
  stable signal; individual episode flips are noise.
- Blind-spot regressions: none -- no task that passed in iter_015 failed in iter_016.
- What was learned: Conservative gating is the key to evidence-summary-memory: only run
  the summary turn when the main-loop answer is unusable, and only adopt the summary
  answer when it is concrete.
- Belief update: `long_context / evidence_noise` is still the only open, harness-fixable
  cell, and it has stable headroom. The remaining placeholder episodes after summary
  suggest the summary prompt is too permissive -- it lets the model emit another
  "unknown" instead of forcing a best guess.
\end{lstlisting}

\subsection{Blind-judge configuration}
\label{sec:app-judgeconfig}

Both judges are run as sandboxed agents, one judgement per candidate pair, in a fresh context,
with no repeats and no self-consistency voting; there is no temperature sweep, and each judge
runs at the same reasoning-effort setting as the predictors. The judge sees one file,
\texttt{judge\_input.md}, containing the mechanically computed per-type deltas as ground truth
and the two predictions anonymized as \texttt{Prediction A} and \texttt{Prediction B}; the
A/B assignment is drawn per candidate from a generator seeded by \texttt{seed + iteration},
and the method-to-label map is stored so the verdict can be decoded afterwards. The judge never
sees which method is which, the method names, or the world models themselves. The cross-provider
replication replays the byte-identical \texttt{judge\_input.md} with the same stored A/B
assignment, so the judge model is the only variable. The instruction is reproduced in full:

\begin{lstlisting}[style=wmc,caption={The judge prompt, verbatim.},label={lst:judge},captionpos=t]
You are an impartial judge comparing the ACCURACY of two predictions. Read
`./judge_input.md` in your working directory. It contains, for ONE proposed scaffold
change:
- the OBJECTIVE realised outcome (which question types actually improved / regressed,
  measured after evaluation -- this is ground truth);
- two independent predictions of that change, anonymised as `Prediction A` and
  `Prediction B`.

Decide which prediction was MORE ACCURATE about reality: it named the question types
that truly moved (gains AND regressions), caught the real regressions rather than
missing them, and got the overall net direction right. Penalise predictions that claimed
improvements that did not happen or missed regressions that did. Judge ACCURACY only --
not writing style or how good the patch was.

Write your verdict to `./judge_verdict.md` with EXACTLY this shape:
WINNER: <A|B|TIE>
REASONING: <2-4 sentences: which types each got right/wrong vs the objective outcome,
and why the winner is more accurate>
\end{lstlisting}

A verdict is counted as a tie when the judge writes \textsc{tie} or when no
\textsc{winner} line can be parsed; ties are excluded from the sign test and reported in
parentheses in Table~\ref{tab:ablation} in the main text.

\section{Initial-Scaffold Stability and Excluded Optimization Diagnostics}
\label{sec:app-stability}

\paragraph{Every memory-benchmark candidate we evaluated held-out.}
Table~\ref{tab:memory-selection} lists them, for both methods, with the train score
that made a candidate eligible and the held-out score it returned. The reported
number in Table~\ref{tab:heldout} in the main text is the best held-out score among each method's
three highest-train candidates, and the table shows what that choice costs: on
LongMemEval-s the eligible spread is $0.5925$--$0.6075$ for \bco and
$0.4825$--$0.5325$ for \nobco, on LoCoMo $0.4451$--$0.4534$ and
$0.3437$--$0.3754$. The selection moves each method by less than the gap between
the methods, but it does select partly on the held-out split and the memory rows
should be read as a best-of-three.

Two candidates were evaluated despite falling outside their method's top three and
are therefore not eligible: \nobco's LoCoMo \texttt{iter015} (train $0.3875$
against a $0.4125$ threshold) scored $0.3796$ held-out, the highest number
either method produced on that benchmark outside the eligible set, and \bco's
LongMemEval-s \texttt{iter014} (train $0.66$ against $0.71$) scored $0.6300$,
the highest either method produced on that benchmark. Quoting either would raise
the corresponding method's number while abandoning the train threshold that makes
the rule a rule, so neither is used.

\begin{table}[t]
  \centering
  {\small
  \setlength{\tabcolsep}{4pt}
  \begin{tabular}{@{}llccc@{}}
    \toprule
    Benchmark & Method & Candidate & Train & Held-out \\
    \midrule
    \multirow{9}{*}{LongMemEval-s}
      & \multirow{3}{*}{\bco}
        & \texttt{iter027} & 0.710 & 0.5925 \\
      & & \texttt{iter025} & 0.690 & \textbf{0.6075} \\
      & & \texttt{iter020} & 0.690 & 0.5950 \\
      \cmidrule(l){2-5}
      & \multirow{6}{*}{\nobco}
        & \texttt{iter030} & 0.590 & 0.5300 \\
      & & \texttt{iter029} & 0.560 & \textbf{0.5325} \\
      & & \texttt{iter025} & 0.560 & 0.4825 \\
      & & \texttt{iter018} & 0.460 & 0.4275 \\
      & & \texttt{iter015} & 0.490 & 0.4000 \\
      & & \texttt{iter012} & 0.450 & 0.3950 \\
    \midrule
    \multirow{8}{*}{LoCoMo}
      & \multirow{3}{*}{\bco}
        & \texttt{iter017} & 0.4750 & \textbf{0.4534} \\
      & & \texttt{iter023} & 0.4625 & 0.4493 \\
      & & \texttt{iter026} & 0.4625 & 0.4451 \\
      \cmidrule(l){2-5}
      & \multirow{5}{*}{\nobco}
        & \texttt{iter013} & 0.4125 & \textbf{0.3754} \\
      & & \texttt{iter016} & 0.4125 & 0.3692 \\
      & & \texttt{iter021} & 0.4125 & 0.3596 \\
      & & \texttt{iter023} & 0.4125 & 0.3437 \\
      & & \texttt{iter015} & 0.3875 & 0.3796 \\
    \bottomrule
  \end{tabular}}
  \caption{Every candidate evaluated on the memory held-out splits, both methods.
  Bold is the value reported in Table~\ref{tab:heldout} in the main text: the best held-out score
  among the method's three highest-train candidates. Rows below each method's train
  threshold were also evaluated and are listed for completeness, but are not
  eligible under that rule. LoCoMo's \nobco train scores are a five-way tie, of
  which four were evaluated.}
  \label{tab:memory-selection}
\end{table}

\paragraph{Spider2-lite.} We repeated the identical initial scaffold four times on the
104-task database-disjoint held-out split.  Aggregate passrates were $0.577$,
$0.519$, $0.558$, and $0.462$.  At the task level, $45/104$ tasks ($43.3\%$)
received both a pass and a fail across the four repeats.  This makes a candidate's
observed delta a weak attribution target: the proposer may revise a belief in
response to a change that would also occur without an edit.

Table~\ref{tab:spider-full} reports every completed selected-candidate evaluation retained
for this diagnostic.  The method ordering changes across campaigns, so we do not
include Spider2 among the five main benchmarks or summarize it with a single
winner.  These runs were exploratory and were not used to define the main
Terminal-Bench split.

\begin{table}[t]
  \centering
  {\small
  \begin{tabular}{@{}lccc@{}}
    \toprule
    Method & Campaign & Train & Held-out \\
    \midrule
    \bco   & 1 & 0.581 & 0.635 \\
    \bco   & 2 & 0.548 & 0.558 \\
    \bco   & 3 & 0.581 & 0.567 \\
    \nobco & 1 & 0.645 & 0.577 \\
    \nobco & 2 & 0.548 & 0.587 \\
    \bottomrule
  \end{tabular}}
  \caption{All retained Spider2-lite selected-candidate evaluations. Train has 31 tasks;
  held-out has 104 database-disjoint tasks. Unequal numbers of completed runs
  are shown rather than silently dropping a campaign.}
  \label{tab:spider-full}
\end{table}

\paragraph{Terminal-Bench split selection.} The recorded dataset is
Terminal-Bench~2.0 (89 tasks), not Terminal-Bench~2.1.  An initial uncurated
train split was dominated by unstable and timeout-bound tasks; its held-out
scores were initial $0.470$, \bco $0.463$, and \nobco $0.455$.  We subsequently
ran repeated initial-scaffold diagnostics and selected a 20-task training subset with more
repeatable failures and usable optimization headroom.  The main table reports
that stability-screened optimization pair.  Because this selection was made
after inspecting initial-scaffold behavior, it remains a potential source of benchmark and
split-selection bias even though it preceded comparison of the optimized methods.

\section{Full Target-Swap Results}
\label{sec:app-transfer-full}

Table~\ref{tab:transfer-full} gives the exact values underlying
Figure~\ref{fig:transfer} in the main text. ``All'' counts a context-window overrun as a
failure; ``common'' restricts every method to tasks all three scaffolds complete. GAIA
has no differential completion, so its two analyses coincide. On AppWorld the
common-set restriction excludes $99$, $77$, and $65$ \bco non-completions at
high, medium, and low effort, respectively.

\begin{table}[t]
  \centering
  {\small
  \setlength{\tabcolsep}{3pt}
  \begin{tabular}{@{}llrccc@{}}
    \toprule
    Benchmark / effort & Set & $N$ & Initial & \nobco & \bco \\
    \midrule
    GAIA / high   & all & 99 & 0.586 & 0.475 & \textbf{0.657} \\
    GAIA / medium & all & 99 & 0.556 & 0.545 & \textbf{0.566} \\
    GAIA / low    & all & 99 & 0.444 & 0.465 & \textbf{0.485} \\
    \midrule
    AppWorld / high   & all    & 372 & 0.556 & \textbf{0.685} & 0.516 \\
    AppWorld / high   & common & 273 & 0.590 & 0.696 & \textbf{0.703} \\
    AppWorld / medium & all    & 372 & 0.500 & \textbf{0.608} & 0.524 \\
    AppWorld / medium & common & 295 & 0.546 & 0.631 & \textbf{0.661} \\
    AppWorld / low    & all    & 372 & 0.438 & \textbf{0.538} & 0.532 \\
    AppWorld / low    & common & 307 & 0.489 & 0.554 & \textbf{0.645} \\
    \bottomrule
  \end{tabular}}
  \caption{Target-model swap to gpt-5.6-luna at three reasoning-effort
  tiers. The three scaffolds are evaluated unchanged. One measurement per
  cell; non-completions count as failures in the full set.}
  \label{tab:transfer-full}
\end{table}

\section{Proposer Token Cost}
\label{sec:app-cost}

\bco's cost profile is not part of the paper's claim, but we report it
because the hypothesis that \bco is additional prompting would predict uniformly higher
proposer token consumption, while a compact-belief account would instead predict lower consumption when
the document reduces re-reading of accumulated evidence. We
report \emph{total} proposer tokens (input $+$ output) relative to \nobco on
matched pairs (Table~\ref{tab:cost}). We quote tokens rather than dollars because
the dollar figure tracks this metric but is provider- and
cache-pricing-dependent; each row compares the two methods of one pair under a
single metering stack, so the Codex-stack Terminal-Bench~2.0 row is internally
consistent even though its absolute token counts are not comparable to the Kimi
rows.

The observed pattern is a read--write trade. \bco reads the raw rollout traces less
often (raw-trace reads down $20$--$69\%$), consistent with the compact append-only belief reducing the need to
re-derive the same information from the bulky history each iteration;
it writes somewhat more, since each iteration emits the updated belief and the
prediction. Where the run's evidence induces heavy re-reading, the read it
saves outweighs the write it adds and the \emph{total} token count falls.

\begin{figure*}[t]
  \centering
  \includegraphics[width=\textwidth]{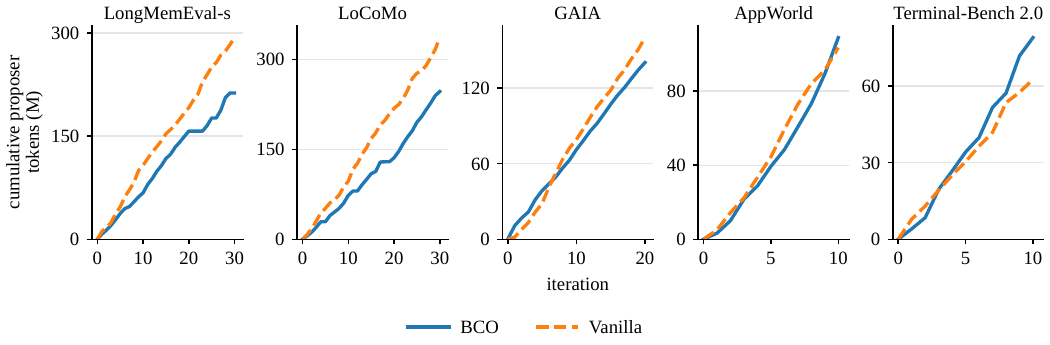}
  \caption{Cumulative proposer tokens vs.\ iteration for \bco and \nobco
  (five matched pairs; Table~\ref{tab:cost}). On the
  memory benchmarks the \bco curve separates below \nobco within the first
  iterations and stays below---the token accounting is consistent with reduced history re-reading from
  early on, not via late amortization. On AppWorld the two lines
  nearly coincide, with \bco crossing slightly above only at the final
  iteration, and on Terminal-Bench~2.0 \bco rises above mid-run: with a few
  bulky trajectories there is little redundant reading to save, while the
  belief's write cost recurs every iteration.}
  \label{fig:tokencurve}
\end{figure*}

The split across benchmarks appears to track the shape of the evidence rather
than the horizon. Totals fall on the memory runs (LongMemEval-s $-27.4\%$, LoCoMo
$-24.7\%$) and on GAIA ($-12.4\%$), and sit at or above parity on AppWorld
($+5.2\%$) and Terminal-Bench~2.0 ($+26.2\%$), even though raw-trace reads
fall everywhere (Figure~\ref{fig:tokencurve}). A within-run control
separates the two explanations for the three runs long enough to admit it:
truncating the memory and GAIA pairs at iteration 10 already reproduces the
split (LongMemEval-s $-37.1\%$, LoCoMo $-24.4\%$, GAIA $-10.0\%$ by iteration
10), so on those pairs the saving is not an amortization effect that accrues
with horizon---it is set within the first iterations by how much redundant
re-reading the benchmark's evidence induces. (The AppWorld and
Terminal-Bench~2.0 pairs run $\horizon{=}10$ to begin with, so the truncation
is vacuous there.) Memory-QA
evidence is many compact per-task artifacts from which \nobco re-derives
its understanding each iteration, and the result is consistent with reduced re-reading
almost immediately; the agentic environments expose a few bulky
trajectories that both methods read once, leaving little redundant reading to
replace while the belief's write cost recurs every iteration. The
Terminal-Bench~2.0 pair, whose metering exposes per-session diagnostics,
makes the decomposition explicit: \bco issues \emph{fewer} tool calls and
file reads per iteration than \nobco yet emits $+45\%$ more output
tokens---the surplus is belief writing, not evidence re-reading. \bco is
thus not uniformly more costly under the additional-prompting hypothesis,
though it is more expensive where history re-reading does not
dominate the proposer's budget.

\begin{table*}[t]
  \centering
  {\small
  \begin{tabular}{@{}lccc@{}}
    \toprule
    Benchmark & $\horizon$ & Proposer tokens $\Delta\%$ & Raw-trace reads $\Delta\%$ \\
    \midrule
    LongMemEval-s        & 30 & $-27.4$ & $-20$ \\
    LoCoMo               & 30 & $-24.7$ & $-69$ \\
    GAIA                 & 20 & $-12.4$ & $-46$ \\
    AppWorld             & 10 & $+5.2$  & $-23$ \\
    Terminal-Bench~2.0   & 10 & $+26.2$ & $-42$ \\
    \bottomrule
  \end{tabular}}
  \caption{Total proposer tokens (input $+$ output) and raw-trace reads, \bco
  relative to \nobco, across matched pairs ($\Delta\%$; negative is fewer).
  Totals fall on the benchmarks whose evidence induces heavy history
  re-reading (memory QA, GAIA) and sit at or above parity on the agentic
  environments whose evidence is a few bulky trajectories; on the three pairs
  with $\horizon>10$ the split is already present when the run is truncated at
  iteration 10 (\S\ref{sec:app-cost}), so on those it is not a horizon effect.
  Each row is a within-pair comparison under one metering stack
  (Terminal-Bench~2.0 uses the Codex stack).}
  \label{tab:cost}
\end{table*}

\section{Persistent Memory Without Calibration}
\label{sec:app-notes-ablation}

The main comparison removes the world-model layer as a whole. To distinguish the calibration
protocol from the simpler benefit of giving the proposer any cross-iteration memory, we run a
three-method diagnostic on LongMemEval-s. \nobco is the comparison method of the main experiments: each
proposer call starts from the raw score, trace, and diff evidence, with no separately
maintained state. The \textbf{Notes} method adds exactly one persistent file,
\texttt{agent\_notes.md}, which the proposer may rewrite or organize however it wishes. It
imposes no prediction, falsifiable belief structure, confidence or status field,
self-grading, or predict--observe--correct update. \bco is the full protocol. The middle method
therefore controls for persistence and summarization without imposing calibration
discipline.

Table~\ref{tab:notes-ablation} reports the best training passrate found in three
independent optimization trajectories per method. Each trajectory has a nominal
ten-iteration proposer budget; the score is taken over all candidates
successfully evaluated within that trajectory. This is an online search
diagnostic on the 100-example training split, not a substitute for the held-out
comparison in Table~\ref{tab:heldout} in the main text.

\textbf{\bco has a modest mean lead, but the diagnostic is noisy.} The
three-run means are $0.430$ (\nobco), $0.427$ (Notes) and $0.467$ (\bco), so
\bco leads by $0.037$ over \nobco and $0.040$ over Notes. However, the
dispersion \emph{within} each method exceeds both gaps. Notes spans $0.39$ to
$0.46$ across its three trajectories: it has the highest score in Runs 1 and
2 but the lowest in Run 3. Three trajectories of a ten-iteration search on 100
training examples are therefore too noisy to establish that Notes is
ineffective or that the calibration protocol reliably outperforms free-form
persistent memory. We report the diagnostic because we ran it, not because it
resolves the question. What does bear on it is elsewhere:
Section~\ref{sec:exp-main} of the main paper is a paired held-out comparison over $400$--$1449$ test
examples, and Section~\ref{sec:exp-ablation} of the main paper isolates the document's \emph{content}
against a control that holds format, length and vocabulary fixed---a cleaner
contrast than one protocol against another, since it varies only the thing in
dispute.

\begin{table}[t]
  \centering
  \small
  \renewcommand{\arraystretch}{1.15}
  \begin{tabular}{@{}lcccc@{}}
    \toprule
    Method & Run 1 & Run 2 & Run 3 & Mean \\
    \midrule
    \nobco & 0.37 & 0.44 & 0.48 & 0.430 \\
    Notes  & \textbf{0.43} & \textbf{0.46} & 0.39 & 0.427 \\
    \bco   & 0.42 & 0.43 & \textbf{0.55} & \textbf{0.467} \\
    \bottomrule
  \end{tabular}
  \caption{Persistent-state ablation on LongMemEval-s. Entries are the
  best-so-far training passrate over successfully evaluated candidates in
  three independent optimization trajectories (100 training examples; nominal
  ten-iteration proposer budget). Notes adds cross-iteration memory but no
  calibration protocol. Each method's within-method range ($0.11$, $0.07$, $0.13$)
  is as large as or larger than any between-method difference in means. Thus,
  although \bco has the highest mean, the diagnostic does not establish that
  Notes is ineffective.}
  \label{tab:notes-ablation}
\end{table}

\section{Per-Run Breakdown of the Offline Ablation}
\label{sec:app-ablation}

Table~\ref{tab:ablation} in the main text aggregates 40 paired candidates. Table~\ref{tab:ablation-perrun}
breaks the same judgements down by run, for both judges. The run is the unit at
which the candidates are actually independent: within a run they share one
world model and one search trajectory. Every one of the three pairings favours
the same method in all four runs, under both judges---twelve run-level
comparisons, twelve in the predicted direction. With four runs, a run-level
sign test cannot fall below $p=0.125$ however consistent the direction is, so
we read this as replication across runs and judges rather than as a small
$p$-value. Per-run win rates are not directly comparable across runs, since
each run contributes a different candidate set sampled at different iterations.

We report downside \emph{precision} rather than recall in
Table~\ref{tab:ablation} in the main text for two reasons. First, recall alone rewards naming
many downside types, which a scrambled document can do without using the original content---demonstrating
that regressions happen is one of the channels scrambling deliberately
preserves. Second, recall is undefined whenever a batch contains no
realized-regressed type at all, which is the case for one of the four runs;
precision is defined for all 40 candidates.

\begin{table}[t]
  \centering
  {\small
  \setlength{\tabcolsep}{4pt}
  \renewcommand{\arraystretch}{1.1}
  \begin{tabular}{@{}llccc@{}}
    \toprule
    Run & Judge & I\,vs.\,N & S\,vs.\,N & I\,vs.\,S \\
    \midrule
    \multirow{2}{*}{LME 1}    & Kimi     & 10--0 & 5--4 (1) & 8--1 (1) \\
                              & DeepSeek & 9--1  & 5--4 (1) & 9--1 \\
    \multirow{2}{*}{LoCoMo 1} & Kimi     & 9--1  & 7--3     & 6--3 (1) \\
                              & DeepSeek & 10--0 & 8--1 (1) & 5--3 (2) \\
    \multirow{2}{*}{LME 2}    & Kimi     & 8--2  & 7--3     & 6--4 \\
                              & DeepSeek & 6--1 (3) & 5--3 (2) & 6--4 \\
    \multirow{2}{*}{LoCoMo 2} & Kimi     & 9--1  & 7--2 (1) & 6--4 \\
                              & DeepSeek & 10--0 & 6--1 (3) & 6--4 \\
    \midrule
    \multirow{2}{*}{Total}    & Kimi     & \textbf{36--4} & \textbf{26--12 (2)} & \textbf{26--12 (2)} \\
                              & DeepSeek & \textbf{35--2 (3)} & \textbf{24--9 (7)} & \textbf{26--12 (2)} \\
    \bottomrule
  \end{tabular}}
  \caption{Blind-judge wins--losses (ties in parentheses) per run, ten
  candidates each, under both judges. I, S and N are the Intact, Scrambled and
  None conditions; LME is LongMemEval-s and the digit is the run. The two judges
  replay identical inputs, including the same A/B position assignment, so the
  judge model is the only difference between the two rows of a run. All twelve
  run-level comparisons favour the same condition; the two judges return the same
  verdict on $97$ of the $120$ pairs.}
  \label{tab:ablation-perrun}
\end{table}

\paragraph{A concrete LongMemEval-s prediction.}
One candidate from LongMemEval-s run~1 replaces tier-priority-first context
ordering with score-first ordering and removes an artificial score boost for
core and summary hits. None and Intact name the same four Upside types but
disagree on the downside. None predicts that ten-word compression will hurt
\texttt{single-session-preference}, which instead improves by $+0.25$. Intact
uses a belief that tier-score miscalibration suppresses evidence and predicts
that removing the boost will let archival or recall hits crowd out
\texttt{single-session-assistant}; that type is the one that regresses, by
$-0.09$. The blind judge therefore selects Intact. This final-document example
illustrates what downside precision measures; it is not additional independent
evidence.

\section{Why Belief Fidelity Is Measured Offline}
\label{sec:app-onlinecurve}

\begin{table}[t]
  \centering
  {\small
  \setlength{\tabcolsep}{4pt}
  \renewcommand{\arraystretch}{1.1}
  \begin{tabular}{@{}lcccc@{}}
    \toprule
    & \multicolumn{2}{c}{LME (run 2)} & \multicolumn{2}{c}{LoCoMo (run 2)} \\
    \cmidrule(lr){2-3}\cmidrule(lr){4-5}
    Per-iteration quantity & 1st & 2nd & 1st & 2nd \\
    \midrule
    Upside hit rate              & 0.464 & 0.244 & 0.267 & 0.133 \\
    Base rate: types improved    & 0.389 & 0.267 & 0.200 & 0.083 \\
    Hit rate $-$ base rate       & 0.076 & $-0.022$ & 0.067 & 0.050 \\
    \midrule
    Types regressed (count)      & 1.6   & 2.6   & 0.9   & 1.5 \\
    Predicted net effect $>0$    & 15/15 & 15/15 & 15/15 & 15/15 \\
    \bottomrule
  \end{tabular}}
  \caption{First and second half of two 30-iteration \bco runs; LME is
  LongMemEval-s. The online upside hit rate falls, but so does the fraction of
  question types that improve at all; the hit rate's excess over that base rate
  falls much less. The predicted net effect is positive in every iteration of
  both runs, so the online score cannot separate a better belief from an easier
  or harder round.}
  \label{tab:online-drift}
\end{table}

A cheaper-looking way to ask whether the belief improves would be to read the
\emph{online} prediction scores---the per-iteration grades the proposer
assigns its own prior prediction---and check whether they rise over a run.
They do not rise, and the reason is that the online score is not an instrument
for belief quality: what it measures drifts systematically over a run, in two
ways we can quantify (Table~\ref{tab:online-drift}).

\paragraph{The prediction is optimistic by construction.} The protocol asks
for one improvement per iteration, so the proposer names Upside types and
stakes a positive net effect essentially always: the predicted net-bet
midpoint is positive in $60/60$ iterations across the two runs whose protocol
version logs per-type scores. A predictor that must bet on improvement every
round cannot raise its score by predicting fewer improvements, only by placing
them better.

\paragraph{The positive class shrinks as the run approaches its ceiling.} The
upside hit rate is a precision-style metric, measured against the set of
question types that actually improved---and that set thins out as the
incumbent gets stronger. The fraction of types realized-improved falls from
$0.389$ to $0.267$ on LongMemEval-s and from $0.200$ to $0.083$ on LoCoMo,
while realized regressions become more frequent ($1.6 \to 2.6$ and $0.9 \to
1.5$ types per iteration). Against a shrinking positive class, a predictor of
constant skill mechanically loses hit rate.

Both effects are visible in the numbers. The raw upside hit rate falls
($0.464 \to 0.244$ and $0.267 \to 0.133$ between the halves of the two runs),
but once the base rate is subtracted the decline is much smaller
($0.076 \to -0.022$ and $0.067 \to 0.050$), and on LoCoMo it is almost
entirely base-rate drift. The mirror image appears in the same logs: downside
recall, whose base rate \emph{grows} as regressions become more common, is
flat or rising ($0.427 \to 0.431$ and $0.682 \to 0.875$). Both curves track
their own base rates more closely than they track anything about the belief,
which is why the paper measures belief fidelity with the offline comparison of
Section~\ref{sec:exp-ablation} of the main paper, where every predictor faces the same fixed
candidates and the base rate is held constant by construction. The
recomputation is reproducible with
\texttt{scripts/analysis/e1\_base\_rate\_drift.py}, which re-derives each
iteration's per-type deltas from the stored score tables and recovers the
logged hit rates exactly.

\end{document}